\documentclass{article} 
\usepackage{array}
\usepackage{colortbl}
\usepackage{iclr2026_conference,times}

\usepackage{amsmath,amsfonts,bm}

\def\eqref#1{equation~\ref{#1}}

\def\1{\bm{1}}

\DeclareMathAlphabet{\mathsfit}{\encodingdefault}{\sfdefault}{m}{sl}
\SetMathAlphabet{\mathsfit}{bold}{\encodingdefault}{\sfdefault}{bx}{n}

\usepackage[backref=page]{hyperref}
\usepackage{url}
\usepackage{graphicx}
\usepackage{booktabs}
\usepackage{subcaption}
\usepackage{amssymb}
\usepackage{wrapfig}
\usepackage{cleveref}

\title{Foundation Model for Cardiac Time Series\\ via Masked Latent Attention}

\author{
  Moritz Vandenhirtz\textsuperscript{1}\thanks{Equal contribution. Correspondence to \texttt{moritz.vandenhirtz@inf.ethz.ch}.}, 
  Samuel Ruiperez-Campillo\textsuperscript{1}\footnotemark[1], 
  Simon Böhi\textsuperscript{2}, 
  Sonia Laguna\textsuperscript{1}, \\
  \textbf{ Irene Cannistraci\textsuperscript{1}, Andrea Agostini\textsuperscript{1}, Ece Ozkan\textsuperscript{2}, Thomas M. Sutter\textsuperscript{1}, Julia E.~Vogt\textsuperscript{1}} \\
  \textsuperscript{1}Department of Computer Science, ETH Zurich, Switzerland \\
  \textsuperscript{2}Department of Biomedical Engineering, University of Basel, Switzerland
}

\iclrfinalcopy 
\begin{document}

\maketitle

\begin{abstract}
Electrocardiograms (ECGs) are among the most widely available clinical signals and play a central role in cardiovascular diagnosis. While recent foundation models (FMs) have shown promise for learning transferable ECG representations, most existing pretraining approaches treat leads as independent channels and fail to explicitly leverage their strong structural redundancy. We introduce the latent attention masked autoencoder (LAMAE) FM that directly exploits this structure by learning cross-lead connection mechanisms during self-supervised pretraining. Our approach models higher-order interactions across leads through latent attention, enabling permutation-invariant aggregation and adaptive weighting of lead-specific representations.
We provide empirical evidence on the Mimic-IV-ECG database that leveraging the cross-lead connection constitutes an effective form of structural supervision, improving representation quality and transferability. Our method shows strong performance in predicting ICD-10 codes, outperforming independent-lead masked modeling and alignment-based baselines.
\end{abstract}

\section{Introduction}

Cardiovascular diseases remain among the leading causes of death worldwide \citep{global2025global}. Clinical diagnosis and monitoring increasingly rely on multimodal data streams including imaging, clinical notes, lab tests, and physiological signals, among which the electrocardiogram (ECG) is the most ubiquitous modality due to its cost-efficiency, non-invasiveness, and mature clinical interpretation pipelines \citep{kashou2023ecg}. Its automated diagnosis has been dominated for decades by expert-crafted features  coupled with classical classifiers \citep{liu2014ecg, chen2018arrhythmia}. Over the last years, deep learning has largely shifted the field toward end-to-end learning from raw ECGs, with convolutional neural networks (CNNs) \citep{ribeiro2020automatic} and recurrent models \citep{ubeyli2010recurrent} as the predominant architectures for many clinical tasks \citep{sau2024artificial, hannun2019cardiologist}.\looseness=-1

More recently, foundation models (FMs) have emerged as a compelling direction to reduce reliance on expensive medical labels and enable transfer across tasks and cohorts \citep{moor2023foundation,tian2024foundation}.
Yet, frontier general-purpose models still lag behind domain experts on clinical benchmarks and remain costly to adapt or deploy in practice \citep{khan2025comprehensive}.
A key reason is that most pretraining pipelines remain largely oblivious to domain structure, particularly in ECG time series, where such structure is particularly explicit.
This structure is not a nuisance; it is an intrinsic self-supervisory signal that modern pretraining objectives rarely exploit directly and that motivates cross-lead representations rather than independent-lead designs.

Self-supervised learning (SSL) offers a scalable alternative to label-heavy supervision in medicine \citep{azizi2021big,moody2025foundation,manduchi2023tree}.
Masked autoencoders (MAEs) \citep{he2022masked} have gained momentum by reconstructing missing content from sparse context, encouraging learning robust, transferable representations. 
In ECG specifically, masked modeling has been explored on an independent-lead basis \citep{na2024guiding} and with language-inspired tokenization schemes \citep{jin2025reading}. Yet, existing methods often tokenize using lead-specific encoders or treat leads as quasi-independent ``channels'', limiting their ability to learn cross-lead correspondence.

Leveraging the coherence of medical datasets, clinical recordings often come as structured multi-view observations that share anatomy and semantics across views. Exploiting this structure via multiview contrast, cross-modal alignment, or multitask learning \citep{laguna2025structure} can improve robustness and label efficiency in multimodal models \citep{mo2024multimed,pellegrini2025radialog,chen2024chexagent}. Recently, \citet{erlacher2025swissbeatsnet} combined multi-lead MAE reconstruction with a lead-alignment objective to enforce cross-lead consistency. 
While effective, such a pairwise alignment comes with limitations \citep{tschannen2023image}, which complicates design and may under-utilize richer, higher-order relationships among leads. 
In contrast, attention-based aggregation is a natural fit for structured latent sets: it supports permutation-invariant processing of variable-size collections while learning which elements are most informative \citep{lee2019set} and has been shown to provide interpretable, instance-weighted summaries in related weakly supervised settings \citep{ilse2018attention}.

In this work, we propose a multi-lead MAE FM that directly capitalizes on ECG structure by learning cross-lead connection mechanisms and modeling higher-order lead interactions by integrating latent attention. Concretely, our contribution is four-fold: (i) we introduce a multi-lead MAE FM with explicit cross-lead connection learning that leverages intrinsic redundancy across leads; (ii) we enhance the FM with latent attention to capture higher-order dependencies beyond pairwise alignment; (iii) we provide empirical arguments for cross-lead connection learning as a scalable form of structural supervision; and (iv) we demonstrate broad clinical and scientific translation spanning coarse ICD-based phenotyping to fine-grained disease classification.

\section{Methods}

\begin{figure}[t] 
    \centering
    \includegraphics[width=0.9\textwidth]{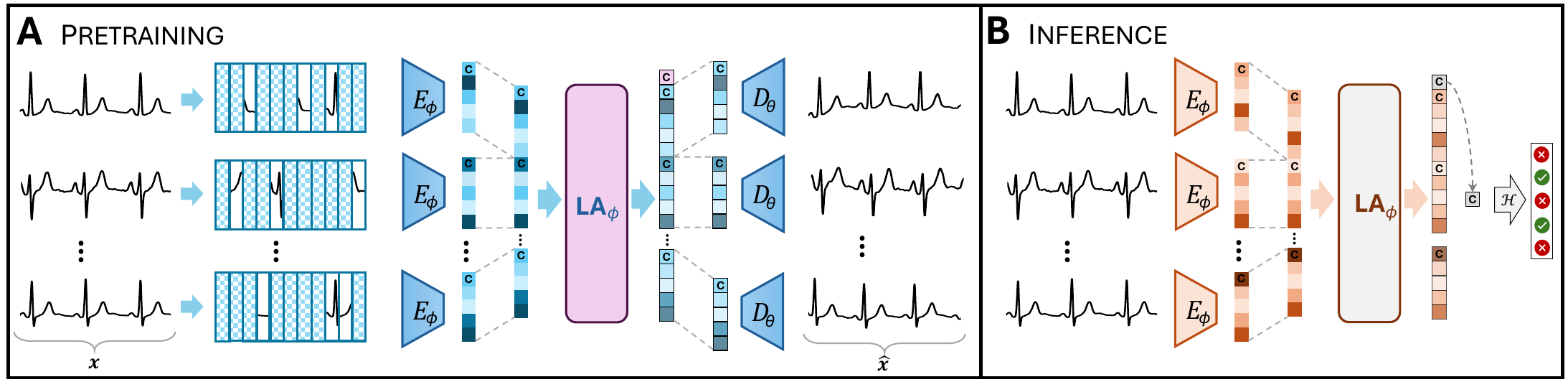}
    \caption{\textbf{Framework overview.} (left) Each ECG is separated into $12$ leads, which are encoded separately and subsequently processed jointly through a latent attention transformer. The training objective is masked reconstruction. (right) The predictions are based on the latent attention's CLS token.\looseness=-1}
    \label{fig:method_pretrain}
    \vspace{-0.5cm}
\end{figure}

We assume an ECG dataset $\mathbb{X} = \{ \bm{X}^{(i)} \}_{i=1}^N$, where $N$ is the number of ECG recordings in the dataset, $\bm{X}^{(i)} = \{ \bm{x}^{(i)}_l \}_{l \in \mathbb{L}}$, $\mathbb{L}$ is the set of leads (e.g. 12 leads in our dataset $\mathbb{L} = \{\ell_I, \ell_{II}, \ell_{III}, \ell_{a_{VR}}, \ell_{a_{VL}}, \ell_{a_{VF}}, \ell_{V_1}, \ell_{V_2}, \ell_{V_3}, \ell_{V_4}, \ell_{V_5}, \ell_{V_6}\}$ for our dataset \citep{strodthoff2024mimic} or see \cref{sec:app_materials}.

The proposed work extends the Masked Autoencoder method \citep{he2022masked} to the ECG domain by introducing a self-attention module in the latent space, i.e., between the encoders $E_\phi$ and the decoders $D_\theta$.
We call the new module \emph{Latent Attention} (LA).

\subsection{Latent Attention for Multi-Lead Integration}
Our latent attention module, inspired by \cite{ilse2018attention,lee2019set}, learns the correlation and shared information between different leads but is flexible enough to not having to merge information between leads in case it would be supoptimal.

We design the latent attention module as a multi-head, multi-layer self-attention block using an additional CLS token similar to the ViT architecture \citep{Kolesnikov2021}.
\begin{align}
    \bm{Z}_{out}^{(i)} = LA_\phi(\bm{Z}_{\text{vis}}^{(i)}) = LA_\phi(M_{\text{LA}}(\bm{Z}^{(i)})) = LA_\phi(M_{\text{LA}}(\{ \bm{z}_l^{(i)} \}_{l \in \mathbb{L}})) 
\end{align}
Similar to encoder $E_\phi$ in the MAE, we apply a random mask $M_{\text{LA}}(\cdot)$ to the input embeddings $\bm{Z}^{(i)}$, i.e., $\bm{Z}_{\text{vis}}^{(i)} = M_{\text{LA}}(\bm{Z}^{(i)})$ with a masking ratio $\alpha_{\text{LA}}$.
See \cref{sec:ecg_lamae} for details on the masking process.\looseness-1


\subsection{ECG-LAMAE FM}
\label{sec:ecg_lamae}

Different to previous works \citep{na2024guiding,jin2025reading}, our FM uses per-lead encoders $E_\phi$ and decoders $D_\theta$ with shared weights $\phi$ and $\theta$.
The latent attention module allows the model to learn the connection between the different leads to extract more meaningful information.

As in the standard MAE implementation, we only feed the visible tokens $\mathcal{T}_{\text{vis}}$ to the encoders $E_\phi$, where $\mathcal{T}_{\text{vis}} \subseteq \{1, \ldots, T\}$ are the indices of the visible patches after applying $M_{\text{E}}(\cdot)$, i.e., $\bm{x}_{l_{\text{vis}}}^{(i)} = M_{\text{E}}(\bm{x}_l^{(i)})$.
We therefore have $T_{\text{vis}} = |\mathcal{T}_{\text{vis}}| = (1 - \alpha_{\text{E}})\cdot T$, where $T$ is the total number of input tokens.

Using our latent attention module, we have the following objective function
\begin{align}
    \mathcal{L} \left(\bm{X}^{(i)} \right) = \frac{1}{\alpha_{\text{E}}} \frac{1}{|\mathbb{L}|} \sum_{l \in \mathbb{L}} \sum_{t \notin \mathcal{T}_{\text{vis}}} \left\| \bm{x}^{(i)}_{l_t} - \hat{\bm{x}}^{(i)}_{l_t} \right\|_2^2, \hspace{0.5cm} \text{where} \hspace{0.5cm} \hat{\bm{x}}^{(i)}_l = D_\phi(LA(\bm{Z}_{\text{vis}}^{(i)})_l)\nonumber 
\end{align}
and $\bm{Z}_{\text{vis}}^{(i)} = M_{\text{LA}}(\{\bm{z}_l^{(i)} \}_{l \in \mathbb{L}})$ is the set of all non-masked latent tokens $\bm{z}_l^{(i)}$ coming from all leads $\bm{x}_l^{(i)}$, i.e., $\bm{z}_l^{(i)} = E_\phi (\bm{x}_l^{(i)})$. This architecture supports the extraction of relevant information of each lead, combined with subsequent merging of the information in the latent attention module for a global representation captured within the CLS token. This token is then used for downstream tasks, as depicted in \cref{fig:method_pretrain} (right).

\section{Experiments and Results}

\begin{table*}[t]
    \centering
    \caption{ICD-10 code prediction performance by hierarchical group under linear probing of the corresponding backbones from the studied models. Best results in \textbf{bold} and second best in \textit{italics}.}
    \includegraphics[width=0.67\textwidth]{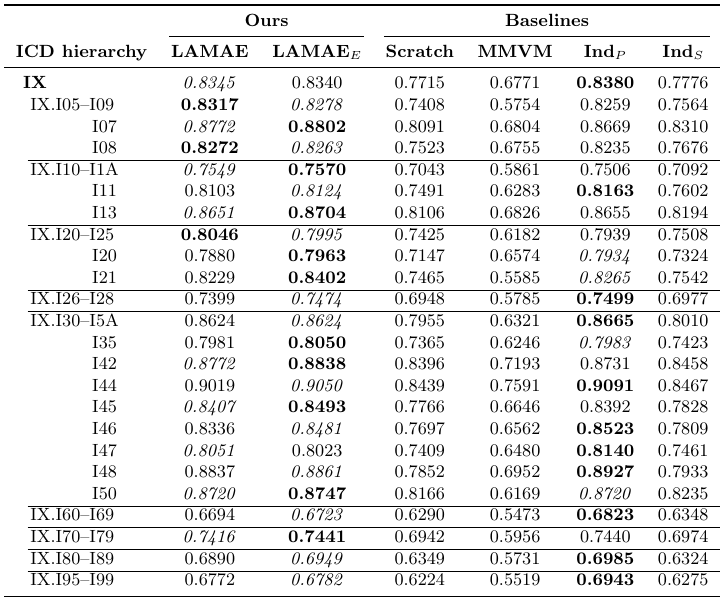}
    \label{tab:icd_hierarchy_results_linear_only}
\end{table*}

We evaluated multi-label ICD-10 prediction from 12-lead ECGs across Chapter IX (I00--I99), spanning valvular disease, hypertensive disease, ischemic syndromes and myocardial infarction, pulmonary circulation disorders, cardiomyopathies, conduction disease, atrial fibrillation/flutter, heart failure, and vascular/cerebrovascular conditions (\cref{tab:icd_hierarchy_results_linear_only} and appendix \cref{sec:app_materials,sec:app_icd,app:extra_results}). Overall, LAMAE-based models achieve strong performance across granularities, with chapter-level AUROC $\approx$0.85 (after fine-tuning; \cref{tab:icd_hierarchy_results_filtered}) and competitive linear-probing results (IX: 0.834; \cref{tab:icd_hierarchy_results_linear_only_app}). Performance is highest for ECG-salient phenotypes, notably conduction and rhythm disorders (e.g., I44 fine-tuning up to 0.9097; I48 up to 0.9016) and acute myocardial infarction subtypes (I21.* often $>$0.93; I210 up to 0.9749), consistent with stereotyped waveform signatures (PR/QRS abnormalities, irregular rhythm, ST/T changes). In contrast, broader vascular and cerebrovascular groupings (I60--I69, I80--I89, I95--I99) are harder from waveform-only inputs (fine-tuning $\sim$0.69--0.72), plausibly reflecting weaker direct ECG imprint and higher label/context heterogeneity. ~\Cref{tab:icd_hierarchy_results_linear_only} is supplemented by an exploration of fine-tuning performance (Tab.~\ref{tab:icd_hierarchy_results_filtered}), which details the greater improvements over linear probing and highlights the benefit of the pretrained backbone for downstream adaptation, as well as the fine-grained hierarchy results for linear probing in Tab.\ref{tab:icd_hierarchy_results_linear_only_app}.\looseness-1

\begin{figure}[t]
\vspace*{-0.4cm}
    \centering
    \includegraphics[width=\textwidth, trim={2cm 0 2cm 0}, clip]{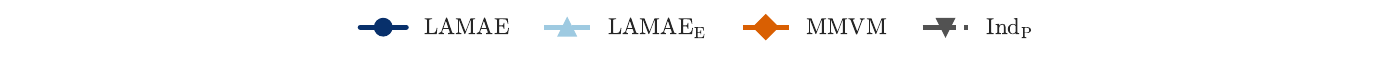}
    \vspace{0.2cm}
    \begin{subfigure}[b]{0.48\textwidth}
         \centering
         \includegraphics[width=\textwidth]{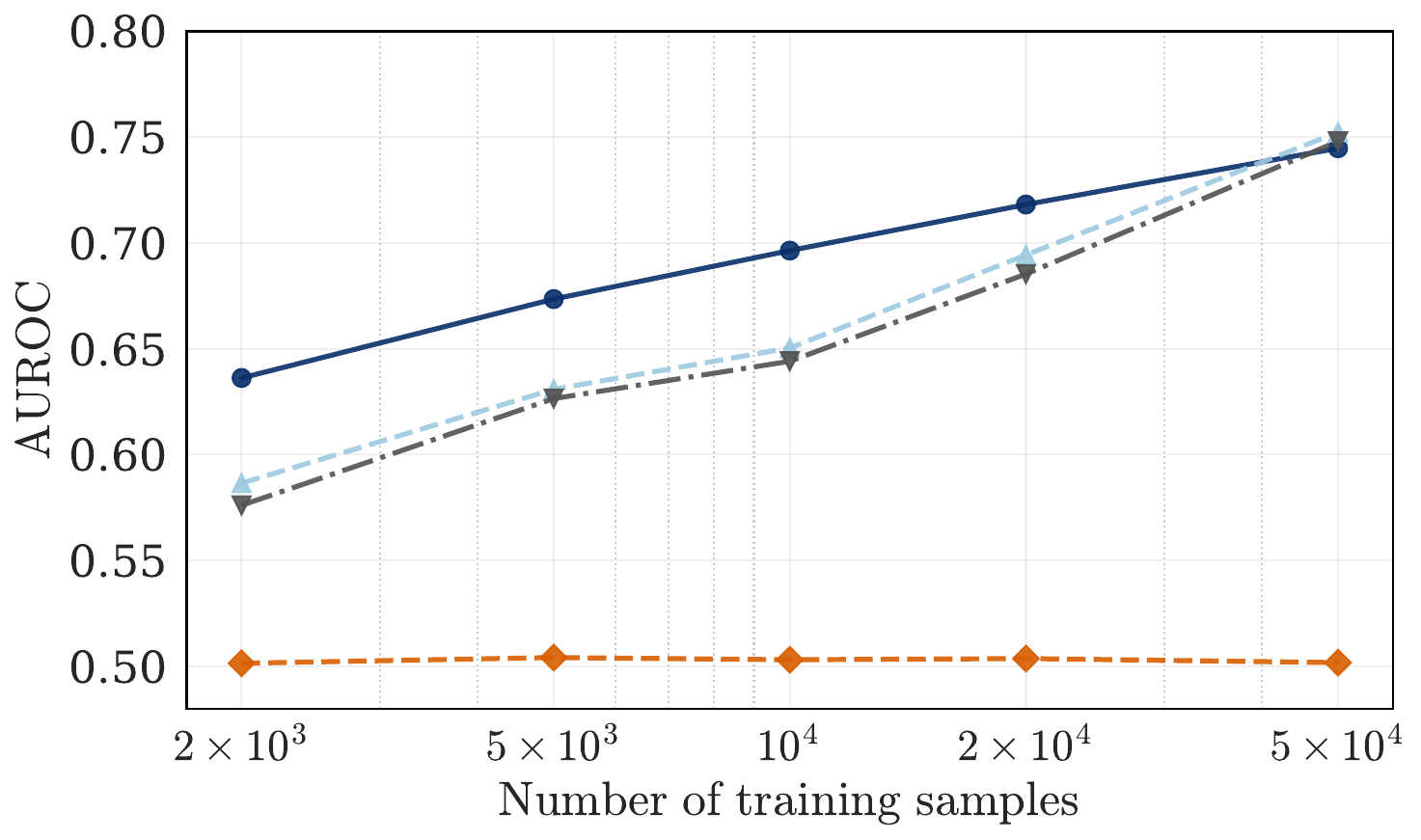}
         \caption{Linear Probing}
         \label{fig:left}
     \end{subfigure}
     \hfill
     \begin{subfigure}[b]{0.48\textwidth}
         \centering
         \includegraphics[width=\textwidth]{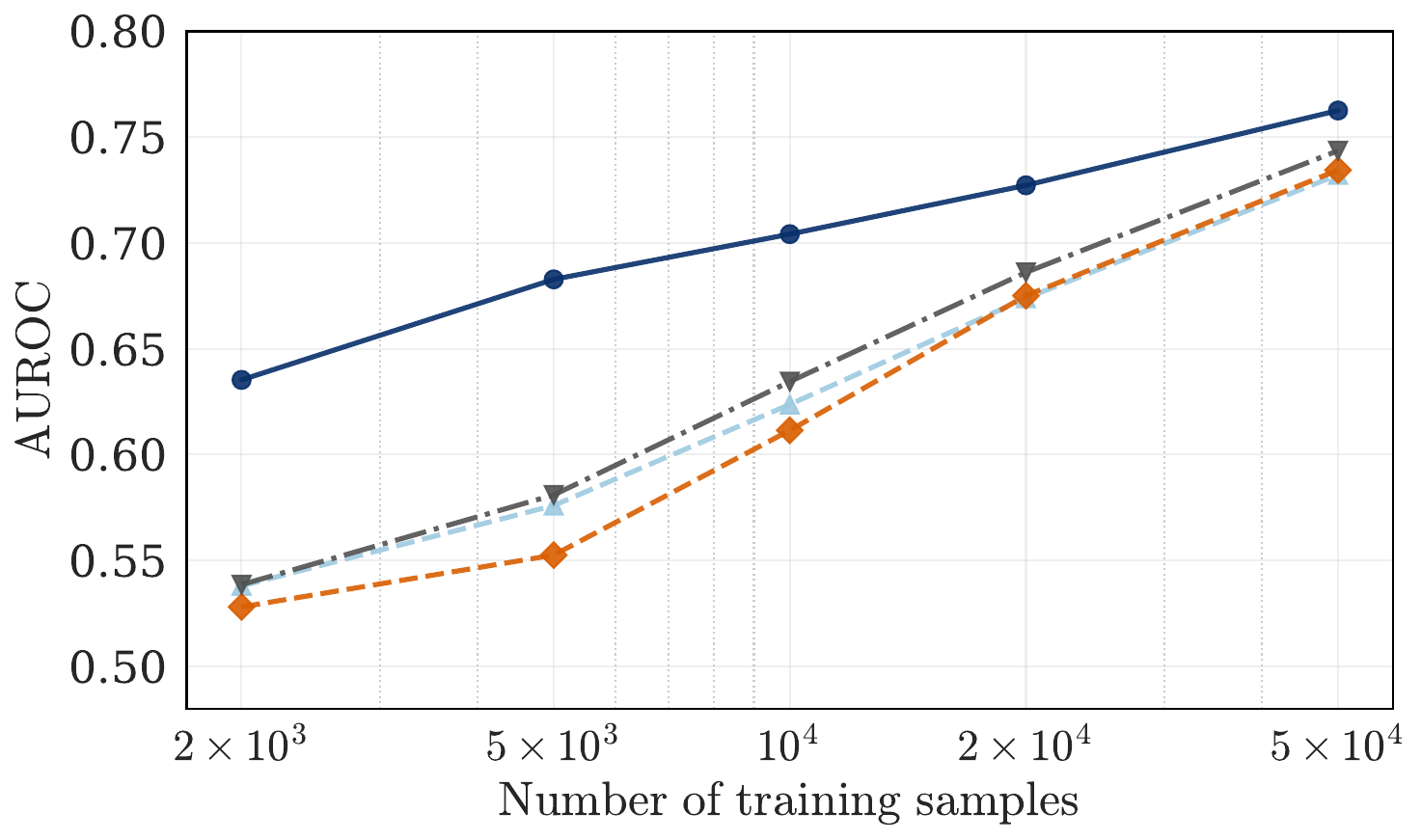}
         \caption{Finetuning}
         \label{fig:right}
     \end{subfigure}
     \vspace*{-0.4cm}
     \caption{\textbf{Label efficiency under finetuning.} Performance curves of the macro-averaged AUROC over all $228$ Chapter IX codes, as a function of the number of training studies used for finetuning.}
     \label{fig:together}
     \vspace*{-0.4cm}
\end{figure}

Scaling experiments show that gains from structure-informed pretraining concentrate in scarce-data regimes (Fig. \ref{fig:together}). Under linear probing and full fine-tuning, LAMAE outperforms scratch-trained and simpler baselines most strongly at small finetuning set sizes, with gaps narrowing only in large regimes (e.g., $\gtrsim$50k samples; Fig. \ref{fig:together}). This supports latent attention as a structure-aware fusion mechanism over correlated lead projections: it can exploit redundancy to encode shared physiology, yielding more sample-efficient representations when curated cardiology datasets are limited.\looseness-1 

A broader ICD-10 benchmarking study is provided by \cite{strodthoff2024prospects}. While not directly comparable, our fine-tuned AUROCs are in a similar range or higher for several overlapping, ECG-identifiable codes, including IX (0.8495), I132 (0.9119), I210 (0.9632), I447 (0.9452), and AF-related subcodes such as I481 (0.8902) and I482 (0.9312) (\cref{tab:icd_hierarchy_results_linear_only_app,tab:icd_hierarchy_results_linear_only_app}). \cite{strodthoff2024prospects} likewise reports strong results for conduction/AF-related codes (e.g., I440 and AF groupings). For the global burden of atrial fibrillation (ICD48 and 48.*; 
\citep{chugh2014worldwide}), our performance is superior to prior task-specific studies that report AUROCs around 0.82--0.85 across external cohorts with CNN-based models \citep{brant2025prediction}, and 0.67--0.8 using demographics or NN-extracted features on a 1-day ECG recording \citep{gadaleta2023prediction}. This is broadly consistent with AF being learnable yet sensitive to cohort shift and label timing. For conduction/heart block phenotypes related to I44 and sub-groups, reported performance varies widely across clinical settings ranging 0.594 to 0.889 \citep{sau2025artificial}, and our results with AUROC above 0.9 suggest that multi-lead structural pretraining can yield robust discrimination even under limited downstream data.

Limitations include the imperfect nature of ICD labels as proxies for physiology and the restricted clinical context available to waveform-only models, particularly for vascular/cerebrovascular diagnoses. Nevertheless, the consistent low-data gains and strong performance on ECG-salient phenotypes indicate that explicitly leveraging cross-lead structure via latent attention is a practical route toward more transferable ECG foundation representations.
     
\section{Conclusion}
We introduced LAMAE: a multi-lead masked autoencoder FM that injects structure into ECG pretraining via latent attention over lead-specific latents. Across a broad Chapter IX ICD-10 hierarchy, LAMAE yields strong AUROC under both linear probing and fine-tuning, with the largest advantages in low-data regimes and in diagnoses where multi-lead interactions are central. These results support that exploiting cross-lead redundancy as structural supervision can improve sample efficiency and downstream transfer, offering a scalable template for time-series foundation models, potentially beyond ECG, where observations naturally come as correlated sets of views.
Even more, latent attention in FMs could serve as a general template for broader applications in science and medicine wherever there is structure between measurements to be leveraged.\looseness-1

\newpage
\section*{Acknowledgements}
This work was supported under project IDs a150 and aa012 as part of the Swiss AI Initiative, through a grant from the ETH Domain and computational resources provided by the Swiss National Supercomputing Centre (CSCS) under the Alps infrastructure. MV and SL are supported by the Swiss State Secretariat for Education, Research, and Innovation (SERI) under contract number MB22.00047.
TS and AA are supported by the grant \#2021-911 of the Strategic Focal Area “Personalized Health and Related Technologies (PHRT)” of the ETH Domain (Swiss Federal Institutes of Technology).
\bibliography{iclr2026_conference}
\bibliographystyle{iclr2026_conference}
\newpage
\appendix
\section{Materials}
\label{sec:app_materials}
We conducted experiments on the MIMIC-IV-ECG-Ext-ICD resource \citep{strodthoff2024mimic}, a PhysioNet \citep{goldberger2000physiobank}release that links raw 12-lead ECG waveforms from MIMIC-IV-ECG \citep{gow2023mimic} to clinically grounded diagnostic labels from the corresponding MIMIC-IV \cite{johnson2023mimic} emergency department and inpatient records. Concretely, ECG acquisition timestamps are aligned with ED stays and hospital admissions to associate each recording with discharge diagnosis codes, providing ICD-10-CM label sets derived from routine clinical documentation rather than retrospective re-annotation. The dataset includes identifiers to retrieve additional clinical context (e.g., ED stay and hospital admission IDs), basic demographics (e.g., age-at-recording, sex), and fold assignments designed to avoid patient overlap for benchmarking and comparability across studies \citep{strodthoff2024mimic}. In our study, only ECG raw waveforms and their paired ICD-10 code were used.

Following the benchmark framing introduced by \cite{strodthoff2024prospects}, we treat ICD-10-CM codes as multi-label targets at multiple granularities (chapter/block/category/subcategory), enabling evaluation from coarse phenotyping to fine-grained diagnosis. Where needed for consistency across label hierarchies, ICD codes may be normalized to a fixed digit format and expanded to include higher-level ancestors in the ICD tree, supporting hierarchical reporting and clinically meaningful aggregation \citep{strodthoff2024mimic,strodthoff2024prospects}.

\section{On ICD codes and further details on those used in this study}
\label{sec:app_icd}
International Classification of Diseases (ICD) codes provide a standardized taxonomy for clinical diagnoses and are routinely used for billing, cohort definition, and large-scale observational research. In this work, we focus on ICD-10 Chapter IX (Diseases of the circulatory system; I00--I99), and report predictive performance at multiple levels of granularity: (i) the chapter-level aggregate, (ii) chapter blocks (e.g., I05--I09), (iii) 3-character categories (e.g., I07), and (iv) selected 4-character subcategories (e.g., I07.1; written as I071). This hierarchical evaluation reflects clinically meaningful groupings while enabling finer assessment of model behaviour on specific diagnoses. An extended description of the clinical meaning per code is included in~\cref{tab:icd_hierarchy_explained}.

\begin{table}[t]
\caption{Clinical meaning of the ICD-10 Chapter IX codes reported in~\cref{tab:icd_hierarchy_results_linear_only_app,tab:icd_hierarchy_results_linear_only}, shown with the same hierarchy.}
\label{tab:icd_hierarchy_explained}
\centering
\includegraphics[width=0.9\textwidth]{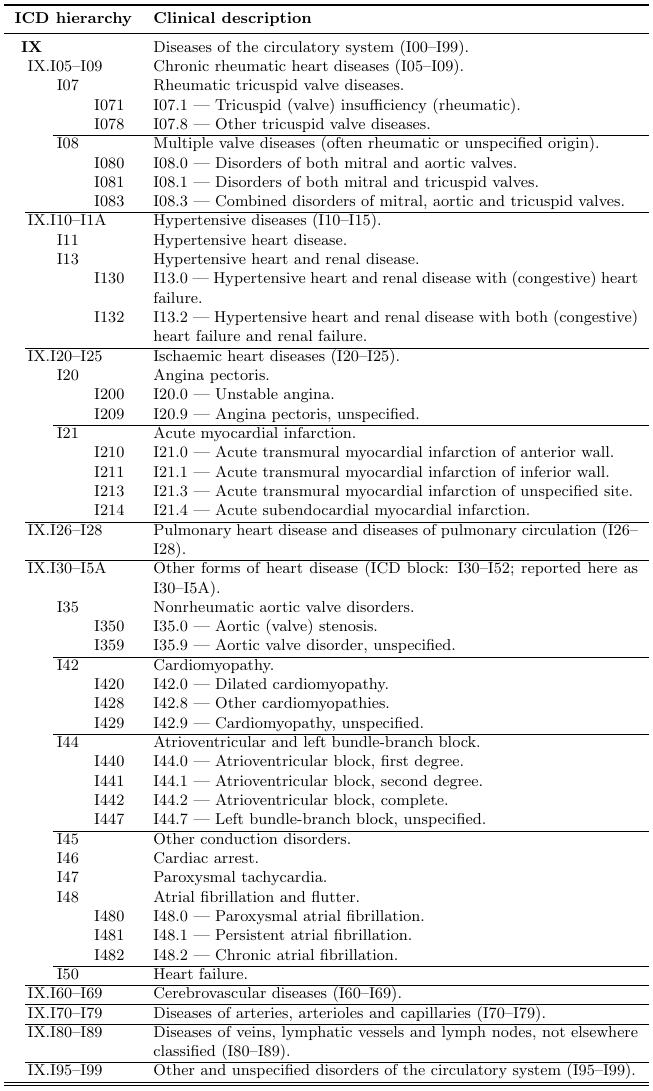}
\end{table}

\section{Supplementary Results: Fine-grained Analysis}

\label{app:extra_results}

\paragraph{Fine-grained Classification Results.} In~\cref{tab:icd_hierarchy_results_linear_only_app}, we present an extended analysis of the fine-grained classification performance initially discussed in~\cref{tab:icd_hierarchy_results_linear_only} on linear probing. The results demonstrate that performance trends remain remarkably consistent across the ICD-10 hierarchy. Notably, the relative advantages of our proposed methods are preserved even as the classification task becomes more granular, confirming the robustness of the learned representations.
\begin{table}[t]
\caption{Extended ICD-10 code prediction performance by hierarchical group in the models studied in~\cref{tab:icd_hierarchy_results_linear_only}. AUROC is reported for linear probing on each corresponding backbone.}
\label{tab:icd_hierarchy_results_linear_only_app}
\centering
\includegraphics[width=0.8\textwidth]{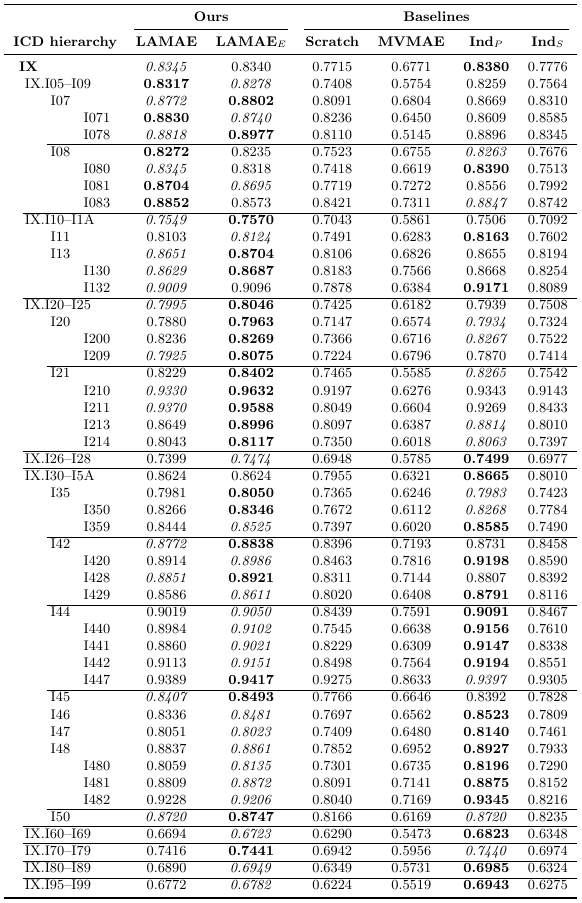}
\end{table}
\paragraph{Performance of LAMAE under Varying Supervision.} Table~\ref{tab:icd_hierarchy_results_filtered} compares the performance of LAMAE and LAMAE$_E$ across both linear probing and fine-tuning regimes. While both models achieve competitive results under linear probing, fine-tuning LAMAE yields a significantly larger performance gain. This suggests that the pretrained backbone serves as a powerful initialization that can be further leveraged to maximize predictive accuracy when labeled data allows for full model updates.
\begin{table}[h!]
\caption{Extended ICD-10 code prediction performance by hierarchical group. Comparison of proposed LAMAE and LAMAE$_E$ under linear probing and fine-tuning of the corresponding pretrained backbone.}
\label{tab:icd_hierarchy_results_filtered}
\centering
\includegraphics[width=0.8\textwidth]{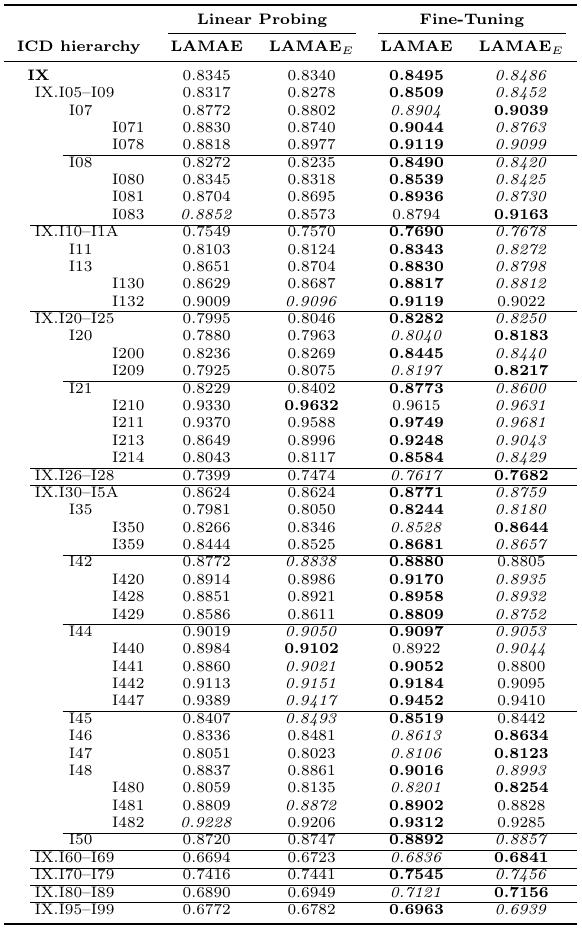}
\end{table}
\end{document}